\documentclass[letterpaper, 10 pt, conference]{ieeeconf}  

\IEEEoverridecommandlockouts                              
\usepackage{cite}
\usepackage{amssymb}
\usepackage{amsmath}
\usepackage[table]{xcolor}
\usepackage{multirow}
\usepackage{graphicx}

\usepackage{caption}
\newcommand{\ptgnet}{\texttt{GPNet\,}}
\newcommand{\hofnet}{\texttt{SRNet\,}}

\newcommand{\robot}{\mathcal{R}}     
\newcommand{\camera}{\mathcal{C}}     
\newcommand{\object}{\mathcal{O}}   
\newcommand{\gripper}{\mathcal{G}}   
\newcommand{\reals}{\mathbb{R}} 
\newcommand{\rbt}[2]{{{^#1} T_{#2}}} 

\newcommand{\Rotf}{\text{Rot}}      
\newcommand{\Transf}{\text{Trans}}  
\newcommand{\graspVector}{t}       
\newcommand{\gtGrasp}{\graspVector^*}       
\newcommand{\nnGrasp}{\hat{\graspVector}}  

\usepackage[normalem]{ulem}

\title{Robotic Grasping through Combined Image-Based Grasp Proposal and 3D Reconstruction}
\author{Daniel Yang, Tarik Tosun, Ben Eisner, Volkan Isler, and Daniel Lee
\thanks{All authors are with the Samsung AI Center NY, 837 Washington St, New York, NY 10014}%
}
\date{October 2020}

\begin{document}

\maketitle

\begin{abstract}
We present a novel approach to robotic grasp planning using both a learned grasp proposal network and a learned 3D shape reconstruction network. Our system generates 6-DOF grasps from a single RGB-D image of the target object, which is provided as input to both networks. By using the geometric reconstruction to refine the the candidate grasp produced by the grasp proposal network, our system is able to accurately grasp both known and unknown objects, even when the grasp location on the object is not visible in the input image.

This paper presents the network architectures, training procedures, and grasp refinement method that comprise our system. Experiments demonstrate the efficacy of our system at grasping both known and unknown objects (91\% success rate in a physical robot environment, 84\% success rate in a simulated environment).  We additionally perform ablation studies that show the benefits of combining a learned grasp proposal with geometric reconstruction for grasping, and also show that our system outperforms several baselines in a grasping task.
\end{abstract}

\section{Introduction}  
Object manipulation in unstructured environments remains a challenging problem in robotics.  Recently, data-driven methods that estimate grasps directly from sensor data (without explicit intermediary state) have resulted in major advances in grasping performance and generalization~\cite{mahler2017dex, zeng2018robotic, mousavian20196, florence2018dense}.  Many of these methods operate on RGB-D image data, and employ convolutional neural network architectures that have proven effective at addressing tasks in computer vision~\cite{he2016deep, huang2017densely}. These systems approach grasping from a perception-driven perspective and often use pixel-space representations of the task, for example, learning affordance maps over pixels \cite{zeng2018robotic} or constraining grasps to the image-plane normal \cite{mahler2017dex}.

While pixel-space representations have clear \textit{computational} advantages, there are clear \textit{physical} advantages to generating full 6-DOF grasps when interacting with real objects.  In addition, in some contexts it is advantageous to infer the presence of a grasp point that is not directly visible in the observed image, either because it represents a favorable affordance for grasping (e.g. the handle of a mug) or because of the constraints of a task (e.g. grasping a visible point might make it difficult to place the object in a desired configuration after grasping).
Many existing image-based data-driven methods formulate grasps by selecting a visible pixel at which to grasp, limiting the grasp plan to a visible point on the object~\cite{mahler2017dex, zeng2018robotic}.

We propose a novel approach to grasp planning using both a learned \textit{grasp proposal} network and a learned \textit{shape reconstruction} network.  
Introducing explicit geometric reconstruction in addition to image-based grasp proposal allows us to fine-tune the proposed grasp for increased accuracy.
Both networks operate on RGB-D images and employ state-of-the-art convolutional architectures, leveraging their representational power and generalization properties.
In contrast to pixel-space methods, our grasp proposal network outputs a full 6-DOF grasp pose, and is capable of proposing grasps on occluded parts of an object not visible in the input image.  Similarly, our 3D reconstruction network can infer the shape of unseen portions of the object. Our system refines the proposed grasp by projecting it onto the reconstructed surface, enabling the robot to precisely grasp both hidden and visible portions of objects. In our experiments, we demonstrate that this refinement step improves grasping performance.

This paper presents the structure of our grasp planning system, including the details of the network architectures, training procedures, and data capture and labelling procedures for the grasp proposal and shape reconstruction networks.  In hardware experiments, we demonstrate that our system is able to successfully grasp instances of an object category (shoes), including cases when the desired grasp point is not visible in the image, and also grasp object instances it was not trained on. In a simulated environment, we further highlight the value in combining the grasp proposal and shape reconstruction networks on another category of objects, boxes from the YCB dataset \cite{ycbdataset2017ijrr}).  Additionally, we include ablation studies to show the benefits of using both shape reconstruction and image-based grasp proposal, especially as the information within the visible portion of the object may be insufficient for refining the predicted grasp, and also show that our system outperforms several baselines systems at the grasping task.
\begin{figure}
    \centering
    \includegraphics[width=\columnwidth]{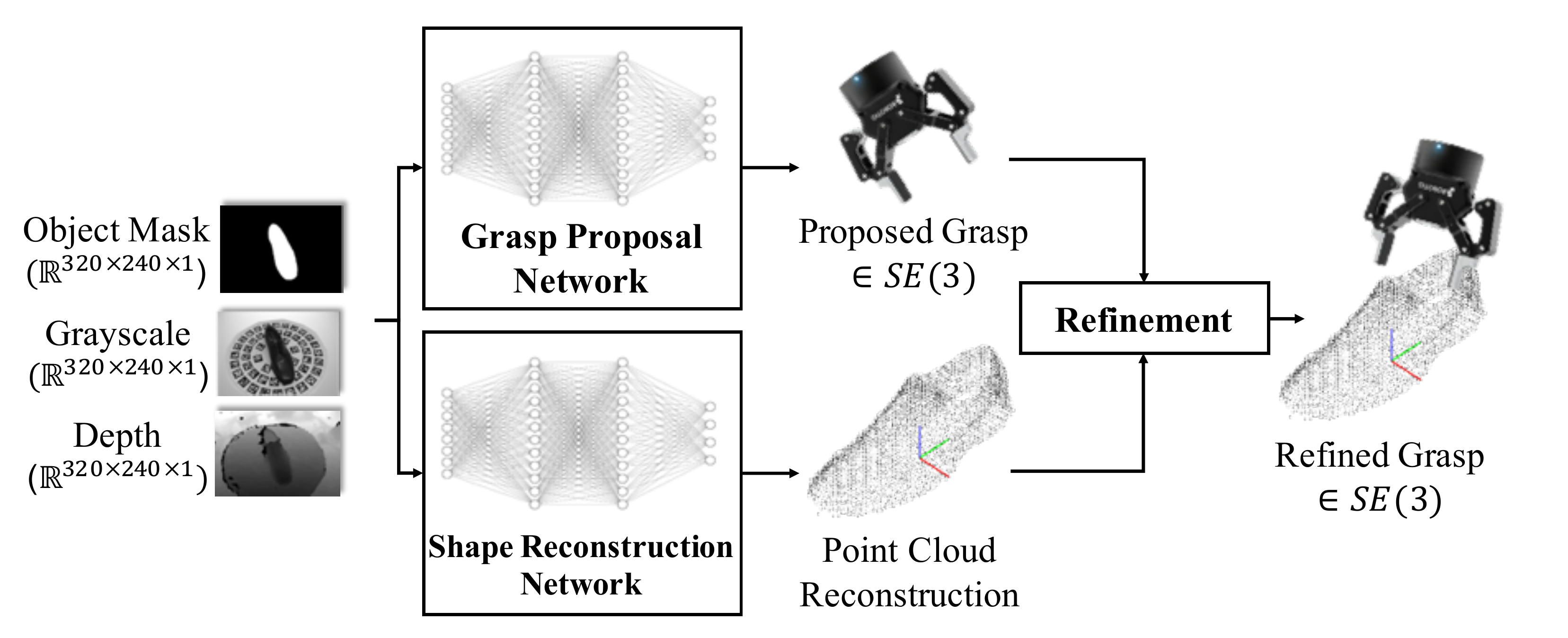}
    \caption{System overview.  An input RGB-D image with segmentation mask is provided as input to two neural networks that produce a 6-DOF grasp pose, and a 3D point cloud reconstruction of the object, respectively.  The grasp pose is refined by projecting it onto nearest point in the point cloud, resulting in the final output grasp.}
    \label{fig:system_overview}
\end{figure}
\section{Related Work}
Robotic grasping is a well-studied problem.  Traditional methods for robotic grasp planning include analytic grasp planners that compute grasp stability metrics using physical models \cite{miller2004graspit}, as well as methods for online object pose estimation that employ optimization techniques to match observed objects to a a library of known models \cite{ciocarlie2014towards}.

Data-driven deep learning methods show promise in improving the efficacy and robustness of robotic grasping.  Many methods operate on RGB-D data, and employ learning paradigms from computer vision.  Some methods learn visual features corresponding to grasp affordances as pixel maps \cite{zeng2018robotic, zeng2019tossingbot}.  Others train networks that evaluate grasp quality from gripper pose and depth image \cite{mahler2017dex, mahler2018dex}.  While these methods have demonstrated good performance at grasping known and unknown objects, the generated grasps are constrained to lie on a visible surface of the object in the image.  Our method generates full 6-DOF grasp poses, and can accurately grasp both hidden and visible portions of objects.

A recent method \cite{atenpas2017gpd} uses geometric heuristics to sample grasp proposals from the visible point cloud and assesses them using a learned quality evaluation network. Another method \cite{mousavian20196} generates grasp proposals by sampling using a variational autoencoder~\cite{kingma2014auto} and, similarly, evaluating these grasps with a learned quality network. \cite{mousavian20196} further refines these grasp proposals with the gradient of their quality network. Our method operates under a similar principle of grasp proposal and refinement, but employs different techniques for both components. For grasp proposal, we perform a single inference step to regress a 6-DOF grasp from an input image using a neural network, whereas \cite{mousavian20196} samples hundreds of grasps using a VAE with a point cloud embedding network and \cite{atenpas2017gpd} is limited to the visible point cloud. For grasp refinement, we project the grasp proposal onto an explicit reconstruction of the target object, whereas \cite{mousavian20196} relies on a grasp quality network that implicitly models the interaction between the gripper and object. We compare our method with \cite{atenpas2017gpd} in our experiments.

Advances from the computer vision and graphics communities have produced methods for general shape reconstruction, which construct an estimated 3D representation of an object from an RGB or RGB-D image of the object \cite{choy20163d, wang2018pixel2mesh, park2019deepsdf, fan2017point}.  Similar methods have been employed to enable robotic grasping \cite{varley2017shape,watkins2019multi}.  These methods use a neural network to reconstruct a 3D model of an object from a depth image, and then plan grasps using an analytic method.  In our work, we employ a recent method that reconstructs an object as a point cloud from a partial-view RGB-D image \cite{mitchell2019higher}.

Category-level grasp planning methods include learned representations using semantic keypoints \cite{manuelli2019kpam}, and representations combining keypoints and shape completion \cite{gao2019kpamsc}.  The primary focus of these efforts is task representation, not grasping; grasps are formulated near keypoints, using a heuristic that accounts for the flatness of the local point cloud and gripper antipodality.  In this paper, we study incorporating 3D reconstruction on data-driven grasp planning.

\section{Problem Statement}
%
%
We consider a setup consisting of a robotic arm with known kinematics, an RGB-D camera, and an object to be grasped. We assume that a method of segmenting an image of the object from its background is available. 

Throughout the paper, $\robot$ denotes the base frame of the manipulator, $\gripper$ denotes the frame of the robot end-effector (a parallel-jaw gripper), $\camera$ is the camera frame and $\object$ is the object body frame. The pose of the camera with respect to the gripper $\rbt{\gripper}{\camera}$ is assumed to be fixed and known, and the pose of the gripper with respect to the robot base $\rbt{\robot}{\gripper}$ is assumed to be computable via forward kinematics. Given a single, segmented RGB-D image of the object from the camera, the objective is for the robot to grasp and pick up the object. Picking is considered successful if at the end of the robot's motion, the object has been lifted completely off the table and is securely grasped by the gripper, and is neither damaged nor deformed by the grasp. In this paper, we train and benchmark our models using shoes as a representative class of objects for our physical setup and with box objects from the YCB dataset for our simulated setup \cite{ycbdataset2017ijrr}.
%
%
\section{Methods} \label{sec:methods}
\subsection{Overview}
Our system (Figure~\ref{fig:system_overview}) is comprised of two networks whose outputs are combined by a refinement module to plan grasps.  The input to our system is a single RGB-D image which is captured, segmented, and provided to both networks.

The \textit{grasp proposal network} \ptgnet outputs a \textit{grasp pose} with respect to the camera frame, $\rbt{\camera}{\gripper} \in SE(3)$. 
The \textit{shape reconstruction network} \hofnet outputs a reconstructed point cloud of the object, providing a reasonable estimate of the shape of the occluded portions of the object.
The outputs of the two networks are combined by projecting the grasp proposal $\rbt{\camera}{\gripper}$ onto the closest point in the reconstructed point cloud, resulting in a refined grasp proposal $\rbt{\camera}{\gripper}^+$.  Since the pose of the camera with respect to the robot $\rbt{\robot}{\camera}$ is known, this camera-frame grasp can be easily transformed into the robot frame for execution by the robot:  ${\rbt{\robot}{\gripper}}^+ = \rbt{\robot}{\camera} {\rbt{\camera}{\gripper}}^+$.

\subsection{Object Dataset}
The dataset used to train \ptgnet for the physical setup is comprised of RGB-D images, masks, and poses of 25 real shoes.  RGB-D images were gathered by placing shoes on a turntable and imaging at 5-degree intervals using three Intel RealSense D435 cameras, for a total of 216 images per shoe.  Images were segmented and fused to form complete point clouds of each shoe, with known camera pose registration from each image to the complete point cloud. The dataset and capture procedure are described in detail in \cite{tosun2019pixels}.

For the simulation evaluation experiments, we use real RGB-D images, masks, and poses of a subset of YCB objects collected using the BigBIRD Object Scanning Rig \cite{ycbdataset2017ijrr, singh2014bigbird}. We specifically focus on the boxed kitchen pantry objects --- the cracker box, sugar box, gelatin box, pudding box, and potted meat can. In contrast to the shoes for our physical setup, these objects are not optimally grabbed along the edge of the object but instead closer to the center of the object which may not always be visible.

\subsection{Grasp Proposal Network - \ptgnet} \label{sec:ptg}
\begin{figure}
    \centering
    \includegraphics[width=\columnwidth]{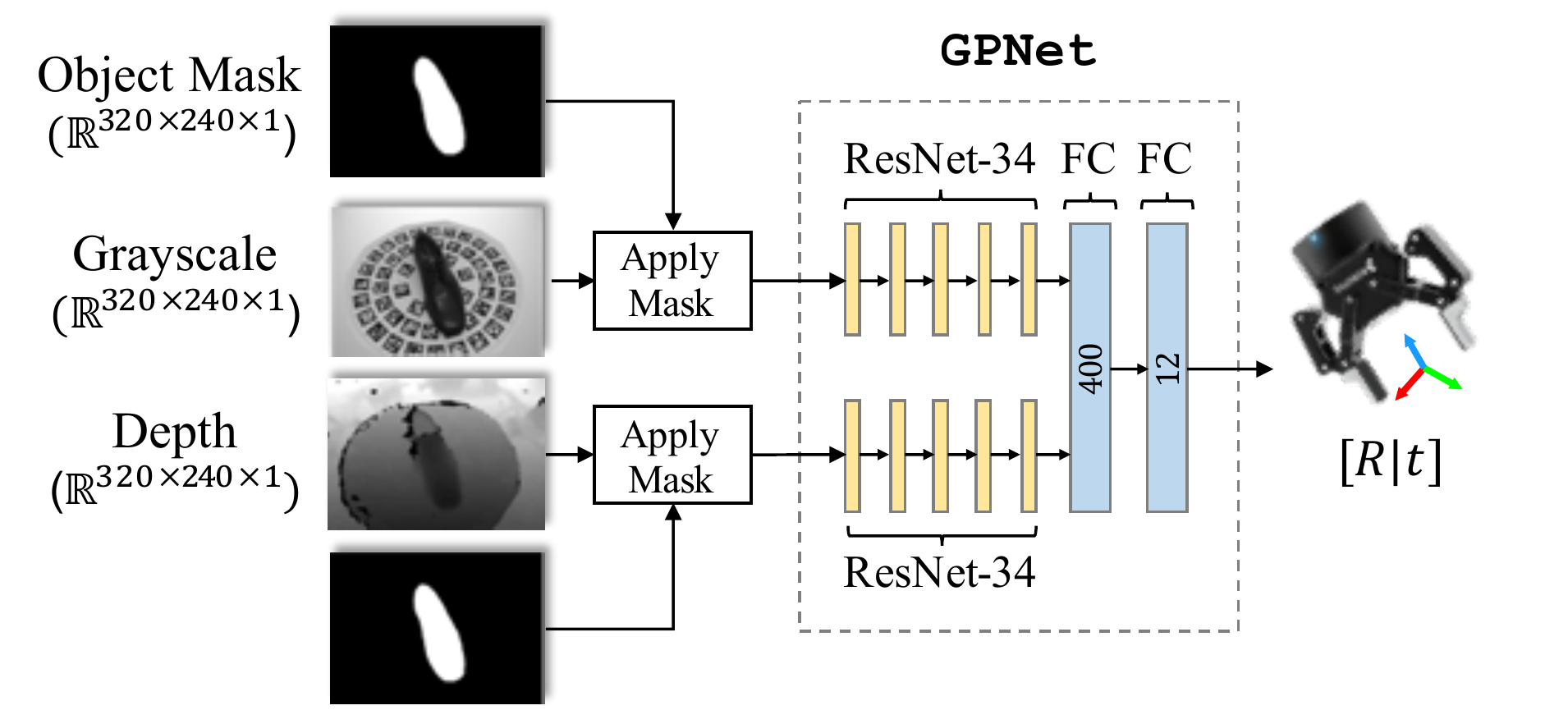}
    \caption{The architecture of \ptgnet consists of parallel ResNet-34 modules that embed the masked grayscale and depth images. These embeddings are concatenated, and regressed via two fully-connected layers into a vector, $\hat{t} \in \reals^{12}$, representing a homogeneous transform, $\hat{\rbt{\camera}{\gripper}}$.}
    \label{fig:ptgnet}
\end{figure}
\subsubsection{Architecture}
\ptgnet (Figure~\ref{fig:ptgnet}) consists of an embedding stage that processes a pair of aligned grayscale and depth images using parallel ResNet-34 convolutional modules \cite{he2016deep}, followed by two fully-connected layers that join the outputs of the RGB and Depth streams, similar in architecture to the network in \cite{tosun2019pixels}.  The input images $(I^g, I^d)$ are assumed to have been foreground masked.  The output of the network is a vector $\hat{t} \in \reals^{12}$ representing a homogeneous transform $\hat{\rbt{\camera}{\gripper}}$, the estimated grasp pose with respect to the camera.  The first 3 values $(t_1, t_2, t_3)$ represent the desired $(x,y,z)$ gripper position while the last 9 values represent a serialized 3D rotation matrix. 
\subsubsection{Example Grasps} \label{sec:example-grasps}
%
\begin{figure}[t]
    \centering
    \includegraphics[width=\columnwidth]{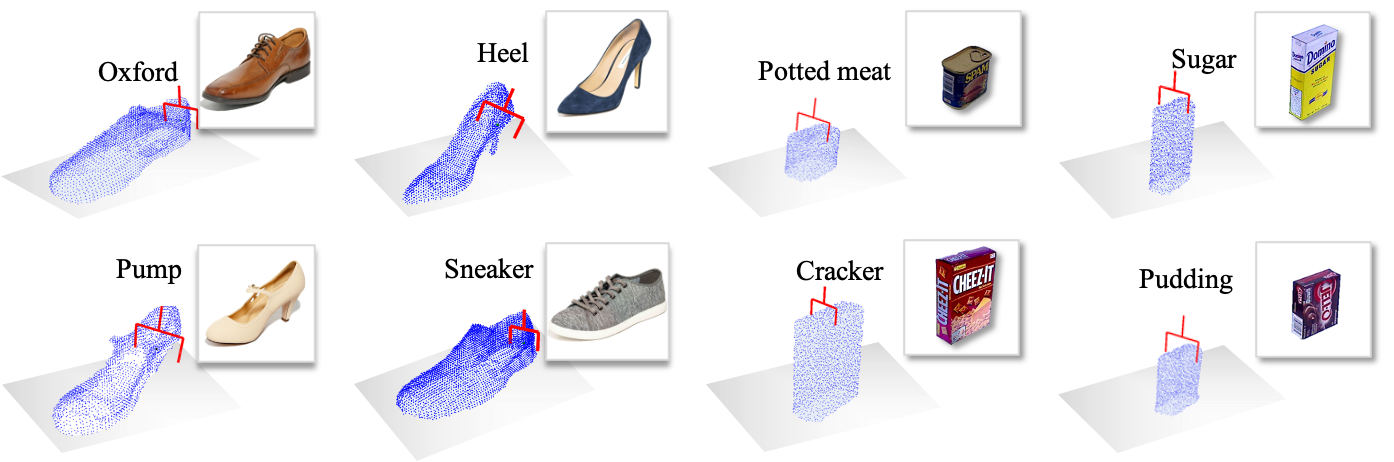}
    \caption{Example grasps in our separate shoe and YCB object datasets. 
    From a set of candidates grasps generated using the 3D meshes, we select a single ground-truth example for each object.}
    \label{fig:shoe_and_ycb_grasps}
\end{figure}



%
Training \ptgnet requires a ground-truth pose of the gripper with respect to the camera (${\rbt{\camera}{\gripper}}^*$) for each RGB-D image in our dataset, which can be calculated knowing the pose of the object with respect to the camera ($\rbt{\camera}{\object}$) and the single object-frame example grasp ($\rbt{\object}{\gripper}^*$) for each \textit{object}.


Ground-truth example grasps are generated for each object using GraspIt, a physics-based grasping simulator~\cite{miller2004graspit}. 
Each object point cloud is converted into a 3D mesh using  marching cubes \cite{lewiner2003efficient}, and provided as input along with a CAD model of the gripper (Robotiq 2F-85). GraspIt samples, evaluates, and optimizes grasps via simulated annealing until the desired number of high-quality grasps have been generated. From these grasps, a single ground-truth example ${\rbt{\object}{\gripper}}^*$ is selected for each object. Selection can be done based on the desired behavior with respect to a task: for example, in the experiments presented in this paper, we manually select grasps as shown in Figure~\ref{fig:shoe_and_ycb_grasps}. For the shoes, we select on the \textit{left} side of the mouth of each example object, while for the YCB objects, we select a top-down grasp towards the object center. For shoes, this might be appropriate for a scenario where the robot needs to grasp and arrange shoes on a shoe store display shelf, with the toe pointed to the left. If a semantically-consistent grasp is not required for a given application, this process could be automated by selecting the grasp with the highest quality score for each object.

%
\subsubsection{Training}
We augment our dataset similar to \cite{tosun2019pixels}, allowing for generalization to non-centered object images. During training, the network is presented with input/output pairs $((I^g, I^d), \gtGrasp)$, where $I^g$ and $I^d$ are foreground masked grayscale and depth images, and $\gtGrasp$ is the corresponding grasp ${\rbt{\camera}{\gripper}}^*$, from the physics-based grasp planner.

Given $(I^g, I^d)$, the network makes a prediction $\nnGrasp$ and a loss is computed from the ground truth grasp $\gtGrasp$.
Let $\Transf(t):\reals^{12} \rightarrow \reals^3$ and $\Rotf(t):\reals^{12} \rightarrow \reals^{3\times 3}$ be functions that extract the position and rotation matrix components of 12-element vector $\graspVector$.
The loss measures the closeness of the predicted and ground truth grasps, and is the weighted sum of a translation and a rotation component:

\vspace{-6pt}
\begin{equation}
    \ell(\nnGrasp, \gtGrasp) = \lambda_T \ell_{T}(\nnGrasp, \gtGrasp) + \lambda_R \ell_{R}(\nnGrasp, \gtGrasp)
\end{equation}

\noindent where $\ell_{T}(\nnGrasp, \gtGrasp)$ is the squared Euclidean distance loss:

\vspace{-6pt}
\begin{equation}
    \ell_{T}(\nnGrasp, \gtGrasp) = ||\Transf(\nnGrasp) - \Transf(\gtGrasp)||^2
\end{equation}

\noindent and $\ell_{R}(\nnGrasp, \gtGrasp)$ is the squared deviation of the predicted rotation times the transpose of the ground truth rotation:

\vspace{-6pt}
\begin{equation}
    \ell_{R}(\nnGrasp, \gtGrasp) = ||\Rotf(\nnGrasp)\Rotf(\gtGrasp)^T - I||^2
\end{equation}

%

\subsection{3D Shape Reconstruction Network - \hofnet} \label{sec:hof}
%
\begin{figure}
    \centering
    \includegraphics[width=0.8\columnwidth]{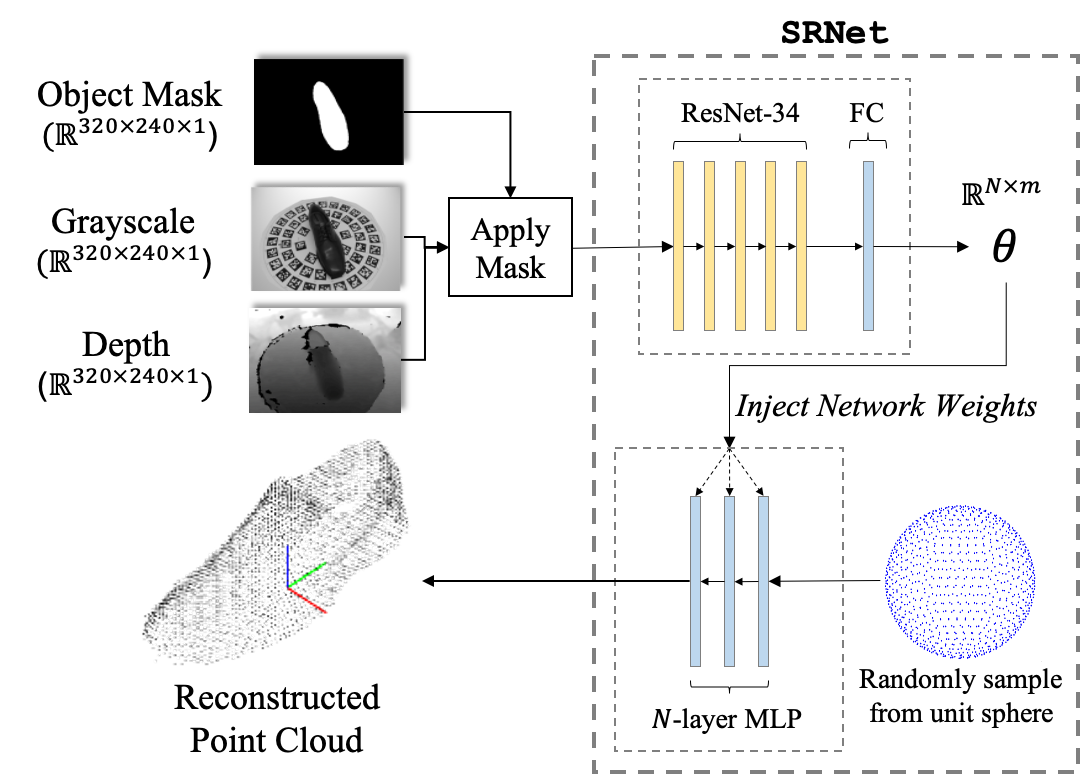}
    \caption{From foreground-masked grayscale and depth images, our 3D shape reconstruction network, \hofnet, learns a mapping function $f_\theta$ that maps points from a canonical domain such as a unit sphere to a 3D object. Our system leverages the additional geometric information provided by this reconstruction to refine grasps proposed by \ptgnet.}
    \label{fig:hof_architecture}
\end{figure}

For the 3D reconstruction component of our system, we use the \hofnet method recently introduced in~\cite{mitchell2019higher}. As outlined in Figure~\ref{fig:hof_architecture}, rather than learning an explicit representation such as a point cloud or an occupancy grid, \hofnet represents an object as a mapping function $f_\theta$ which maps points from a canonical domain $S$ (e.g. a unit sphere in $\reals^3$) to the object surface. The mapping function is represented as a Multi Layer Perceptron (MLP) whose weights $\theta$ are output from a \emph{Higher Order Function (HOF)}. In other words, the higher order function  takes an image as input and outputs the mapping function from $S$ to the object.

Let $I$ be the segmented image of an object and $Y^*$ be the ground truth represented as a set of points sampled from the 3D model of the object. The network is trained as follows: on input $I$, the higher order function outputs $\theta = g(I)$ which corresponds to the weights of the MLP representing $f_\theta$. Afterwards, a fixed number of points $X = \{ x_i \}$ are sampled from the canonical domain uniformly at random. The network output $Y$ is obtained by applying $f_\theta$ on each $x_i$ to obtain $Y = \{f(x) : x \in X\}$.
For training, the Chamfer distance between $Y$ and $Y^*$ is used as a loss function. The advantage of this formulation is that since the domain is resampled at each iteration, the network can learn representations in arbitrary resolution determined only by the ground truth resolution. This representation is comparable to the state of the art in terms of reconstruction accuracy using significantly fewer parameters~\cite{mitchell2019higher}. To align the generated 3D reconstruction with the object, we use iterative closest point (ICP) \cite{rusinkiewicz2001efficient} as implemented in Open3D \cite{Zhou2018}. 
\subsection{Grasp Refinement And Execution} \label{sec:snapping}
\subsubsection{Grasp Refinement}
\begin{figure}
    \centering
    \includegraphics[width=0.3\columnwidth]{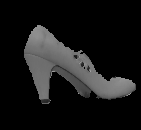}
    \includegraphics[width=0.3\columnwidth]{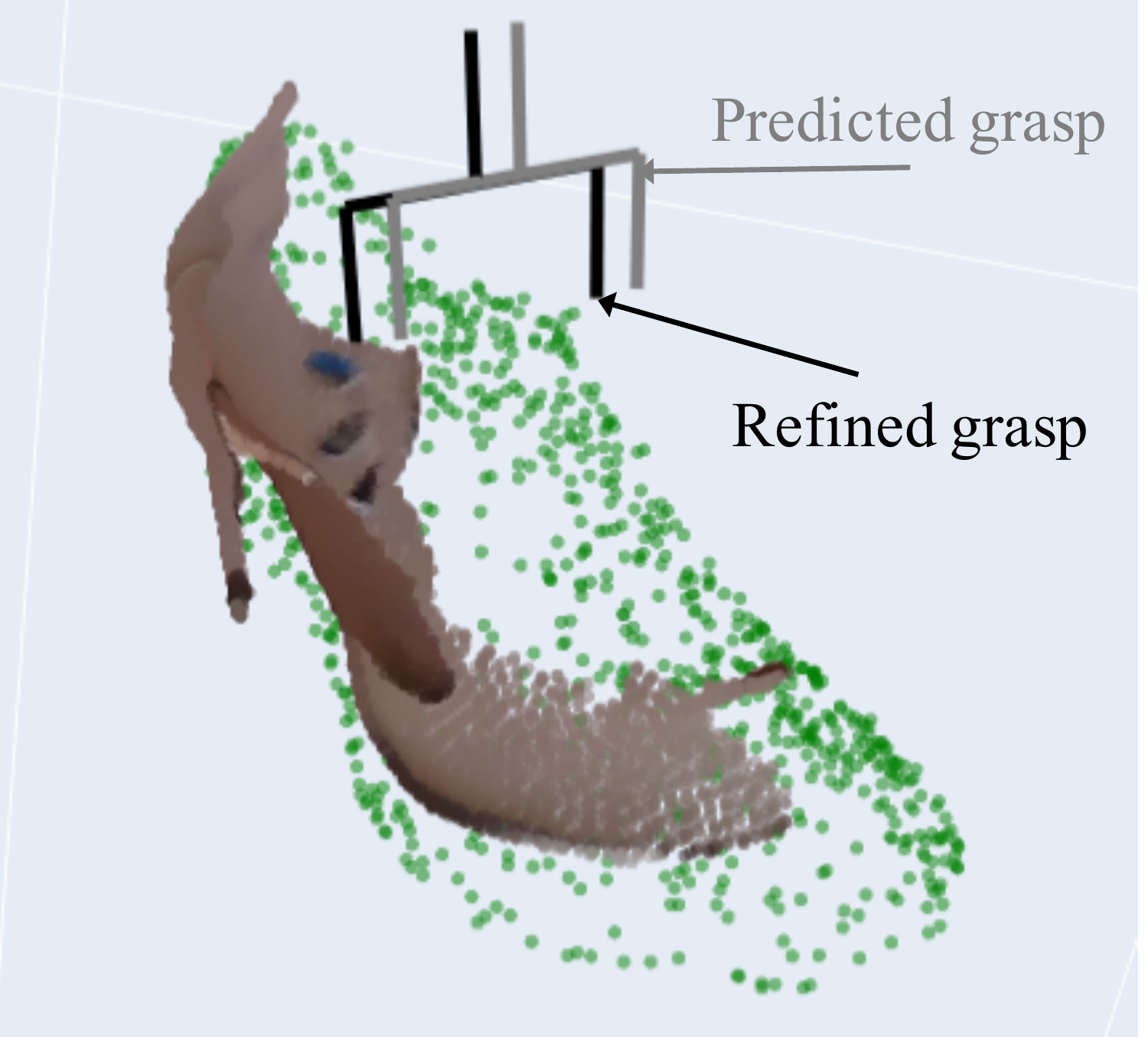}
    \caption{Projecting the proposed grasp onto the reconstructed point cloud improves grasp accuracy.  Left: Grayscale shoe image.  Right: \hofnet reconstruction overlaid on visible point cloud}
    \label{fig:snapping}
\end{figure}
As discussed in Section~\ref{sec:example-grasps}, \ptgnet produces grasp proposals semantically consistent with the training examples. For the network used in our experiments, it will always propose grasps on the left side of the shoe, even if only the right side is visible in the image, as in Figure~\ref{fig:snapping}.  In these cases, \ptgnet must ``guess'' an appropriate location for the grasp, and its accuracy tends to be lower than when grasping a visible point. 

Our system refines the grasp proposed by \ptgnet by projecting it onto the surface of the point cloud reconstruction created by \hofnet.  
Given the proposed grasp $\rbt{\camera}{\gripper}$, we select the point in the point cloud that is closest to the grasp point and refine the grasp by setting its position to the coordinates of that point, leaving its orientation unchanged, resulting in the output grasp $\rbt{\camera}{\gripper}^+$.  Figure~\ref{fig:snapping} provides an illustrative example.
Refining the grasp through this projection operation increases the accuracy of the overall system, and results in better performance in our grasping experiments.  
\subsubsection{Grasp Execution}
Our system produces grasp plans in the form of gripper poses with respect to the camera frame.  Since we assume the pose of the camera $\rbt{\robot}{\camera}$ is known, this camera-frame grasp can be easily transformed into the robot frame:  $\rbt{\robot}{\gripper}^+ = \rbt{\robot}{\camera} \rbt{\camera}{\gripper}^+$.  When executing a grasp on the robot, we command the gripper to first move to an offset point $20cm$ back from the grasp point (along the gripper axis), then move in to the grasp point and close its fingers.
%
\section{Experiments}
We first perform experiments to characterize the performance of our method at grasping known and unknown shoes. 
All systems tested assume a single shoe placed on a flat table, imaged by a depth camera with a known pose with respect to the table. Shoes are segmented from the background by reprojecting the depth image into a point cloud in the table frame, and then removing all visible points beneath the table plane.  We benchmark the network our full system, ablations, and baselines on the set of four shoes shown in Figure~\ref{fig:shoe_and_ycb_grasps}.  One of them (the Oxford) was withheld from the training data for \ptgnet and \hofnet as a test set shoe.

To further highlight our method using a more general set of objects, we perform experiments using objects from the YCB Dataset~\cite{ycbdataset2017ijrr, singh2014bigbird} using real images but simulated grasps.
 
\subsection{Systems Tested}
We compare our full system against seven alternatives, which we subdivide into \textit{baseline}, \textit{ablation}, and \textit{oracle} systems.
Each tested system consists of a grasp proposal and a grasp refinement method, as described below.
\subsubsection{Grasp Proposal Methods}~\\
\textbf{\ptgnet (ours)}
Grasp proposal network in Section~\ref{sec:ptg}.\\
\textbf{Naive (baseline)}
This method proposes an initial grasp 20 cm above the table at the shoe centroid with a uniform random offset of $\pm2$ cm in the table plane to perturb the grasp off-center. The gripper is pointed down, and its grip axis is aligned to the major axis of the visible point cloud.
This is usually not a reasonable grasp on its own, so success depends largely on the projected point on the estimated surface. \\ 
\textbf{Library (oracle)}
This method proposes the ground-truth training-set grasp ${\rbt{\object}{\gripper}}^*$ for the object being tested. Object pose with respect to the camera ${\rbt{\camera}{\object}}$ is estimated based on the available point cloud to move this into camera frame.
\subsubsection{Grasp Refinement Methods}~\\
\textbf{\hofnet (ours)}
Projection onto the output of the 3D reconstruction network in Section~\ref{sec:hof}.\\
\textbf{Visible (baseline)}
Projection onto the visible point cloud.\\
\textbf{None (baseline)}
No refinement step.\\
\textbf{Library (oracle)}
Projection onto the ground-truth point cloud for the object, providing an upper bound on the performance of shape reconstruction. 
\subsubsection{Combinations Tested}~\\
\textbf{Full System (ours)}
Our full system employs \ptgnet for proposal and \hofnet for refinement, as described in Section~\ref{sec:methods}.\\
\textbf{Ablation Systems}
The ablation systems test the effect of replacing either the proposal or refinement stages of our full system with the baseline alternative.  \ptgnet-Visible projects the \ptgnet proposal onto the visible point cloud, and Naive-\hofnet projects the naive grasp proposal onto the \hofnet-generated point cloud. The pose of the Naive grasp proposal is based off of the alignment of the \hofnet point cloud to the visible point cloud. \ptgnet-None tests the performance of the \ptgnet grasp proposal on its own, with no refinement.\\
\textbf{Baseline Systems}
The baseline systems refine the Naive and Library grasp proposals with the Visible point cloud. \\
%
\textbf{Oracle Systems}
The oracle systems test the effect of replacing either the proposal or refinement stages of our full system with the ground-truth alternative, and represent an upper-bound on performance.  
\subsection{Experiment Procedures}
%
\begin{figure}
    \centering
    \includegraphics[width=0.4\columnwidth]{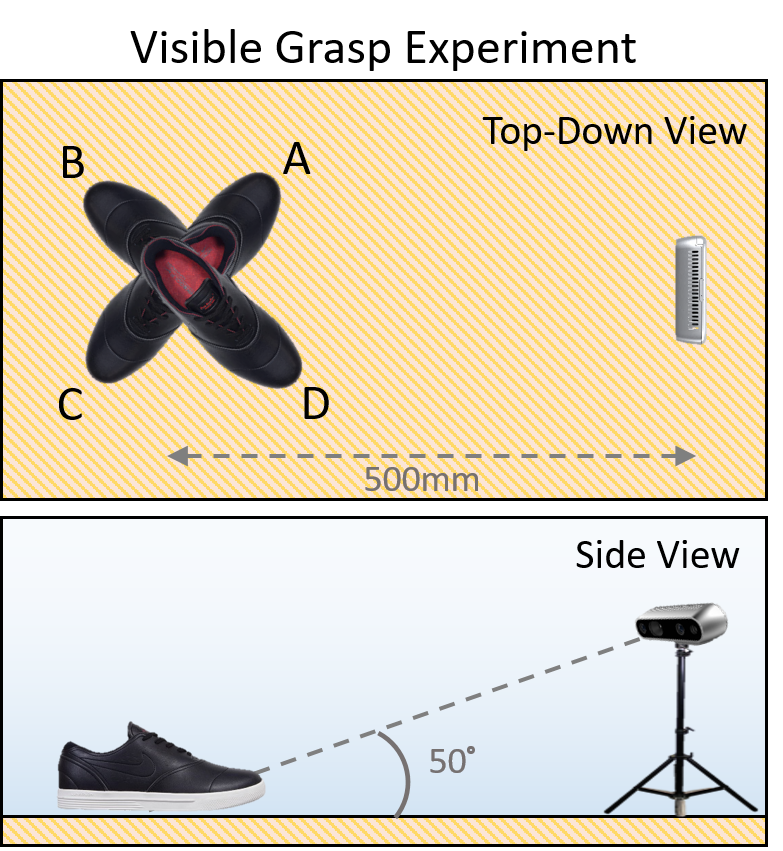}
        \includegraphics[width=0.4\columnwidth]{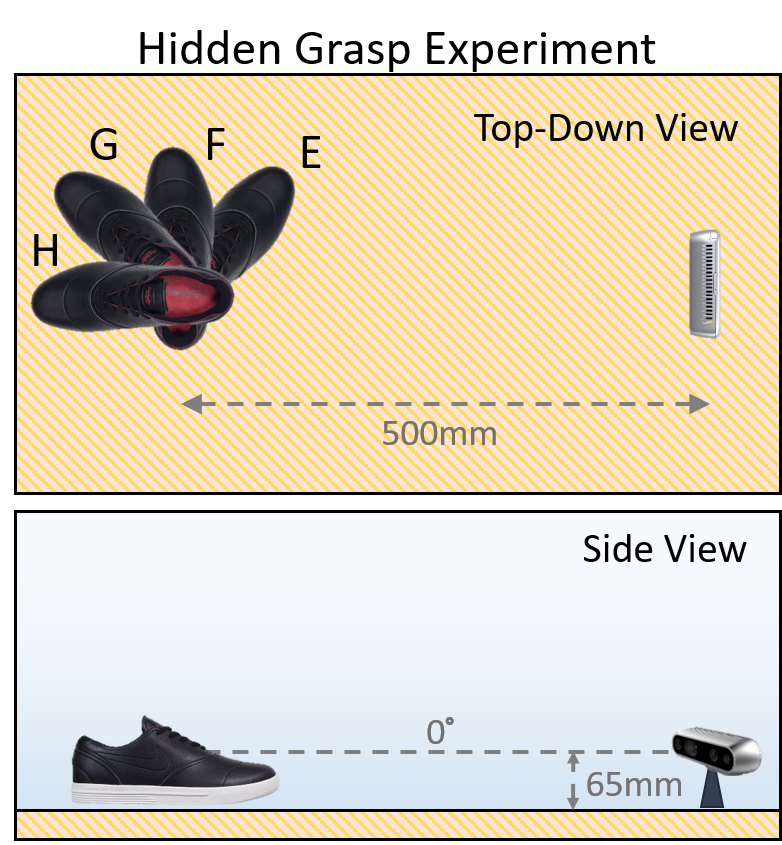}
    \caption{Visible-grasp (left) and hidden-grasp (right) experimental setups.  In the visible-grasp experiment, the camera is $500mm$ away from the shoe at a $50^\circ$ elevation angle, and the shoe is placed in one of four poses (A-D) at $90^\circ$ increments.  In the hidden-grasp experiment, the camera is $500mm$ away from the shoe and parallel to the table, while the shoe is placed in one of four poses (E-H) at $45^\circ$ increments.  The grasp point on the left side of the shoe is not visible from these viewpoints.}
    \label{fig:experimental_setup}
\end{figure}
\subsubsection{Experiment 1: Visible-Grasp Shoe Picking Task}
This experiment tests the ability of each system to pick up shoes from camera viewpoints that provide an unobstructed view of the grasp point. Each shoe in Figure~\ref{fig:shoe_and_ycb_grasps} was tested.  The Heel, Pump, and Sneaker were all included in the training set, while the Oxford was withheld as a test set shoe.

Figure~\ref{fig:experimental_setup} shows the experimental setup. In each trial, the shoe is placed on the table, in one of four orientations at $90^\circ$ increments (labelled A, B, C, and D).  The camera is then positioned to point at the shoe, at a distance of $500$ mm
and with an elevation angle of $50^\circ$ (from the table plane). 

%
\subsubsection{Experiment 2: Hidden-Grasp Shoe Picking Task}
This experiment tests shoe picking from viewpoints where the desired grasp point is occluded in the visible image.  The camera and shoe are positioned as shown in Figure~\ref{fig:experimental_setup}: the camera is parallel to the table ($0^\circ$ elevation) at a height of $65$ mm and a distance of $500$ mm to the shoe, which is placed in one of four poses at $45^\circ$ increments (labelled E, F, G, and H).  In each of these poses, the grasp point (on the left side of the shoe) is hidden from view.  For example, Figure~\ref{fig:snapping} shows an image of the Pump taken in pose G.
\subsection{Results}

%
\begin{table}
    \centering
\begin{tabular}{|c|cc|cc|cc|}
\hline
Type & Proposal & Refinement & VG & \% & HG & \% \\ \hline
Baseline & Naive & Visible & 6/16 & 38 & 3/16 & 19\\
Baseline & Library & Visible & 8/16 & 50 & 6/16 & 38\\ \hline
Ablation & Naive & \hofnet & 6/16 & 38 & 8/16 & 50\\
Ablation & \ptgnet & None & 14/16 & 88 & 12/16 & 75\\
Ablation & \ptgnet & Visible & 15/16 & 94 & 9/16 & 56\\ \hline
\bf{Full} & \bf{\ptgnet} & \bf{\hofnet} & \bf{15/16} & \bf{94} & \bf{14/16} & \bf{88}\\ \hline
Oracle & \ptgnet & Library & 14/16 & 88 & 15/16 & 94\\
Oracle & Library & \hofnet & 13/16 & 81 & 16/16 & 100\\ \hline
\end{tabular}
    \caption{Shoe-Picking Experiment Success Rates for Visible-Grasp (VG) and Hidden-Grasp (HG) settings }
    \label{tab:new_grasp_experiment}
\end{table}

%
%

\begin{table}
    \centering
\setlength{\tabcolsep}{5pt} 
\begin{tabular}{|c|cc|cccc|cc|}
\hline
Type     & Prop. & Ref. & \bf{O}    & H     & P     & S     & Total     &   \% \\ \hline
Baseline & Naive    & Visible    & 1/8  & 4/8   & 2/8   & 2/8   & 9/32      & 28 \\
Baseline & Library  & Visible    & 2/8  & 3/8   & 7/8   & 2/8   & 14/32     & 44 \\ \hline
Ablation & Naive    & \hofnet        & 1/8  & 5/8   & 4/8   & 4/8   & 14/32     & 44 \\ 
Ablation & \ptgnet      & None       & 8/8  & 7/8   & 6/8   & 5/8   & 26/32     & 81 \\ 
Ablation & \ptgnet      & Visible    & 6/8  & 6/8   & 7/8   & 5/8   & 24/32     & 75 \\ \hline
\bf{Full}& \bf{\ptgnet} & \bf{\hofnet}   & \bf{7/8}  & \bf{7/8}   & \bf{8/8}   & \bf{7/8}   & \bf{29/32}     & \bf{91} \\ \hline
Oracle   & \ptgnet      & Library    & 7/8  & 7/8   & 8/8   & 7/8   & 29/32     & 91 \\ 
Oracle   & Library  & \hofnet        & 8/8  & 5/8   & 8/8   & 8/8   & 29/32     & 91 \\ \hline


\end{tabular}
    \caption{Per-shoe grasping success rates from both viewpoints. O=Oxford, H=Heel, P=Pump, S=Sneaker.  Oxford is a test shoe.}
    \label{tab:per-shoe-results}
\end{table}
Table~\ref{tab:new_grasp_experiment} show the results for the visible- and hidden-grasp experiments.  Our combined system (Full \ptgnet-\hofnet) performs well at the grasping task under both conditions, with a 94\% success rate when the grasp point is visible, and an 88\% success rate when the grasp point is hidden.  We see a clear performance advantage compared to the baseline systems under both conditions, which have task success rates of under 50\%.  Comparing the full system to the oracle systems (which represent an upper bound on performance), we see that its performance is comparable in both cases.

In the visible-grasp experiment, the performance of our full system is equal to that of our proposal-only ablation (\ptgnet-Visible), at a 94\% success rate.  This makes sense: when the grasp point is visible in the image, refining the grasp proposal by projecting it onto the visible point cloud should provide the same benefit as projecting it onto the reconstructed point cloud.
In the hidden-grasp experiment, we see that 3D reconstruction provides a clear advantage: our full system succeeds in 88\% of trials, while both ablations (\ptgnet-Visible and Naive-\hofnet) succeed at around 50\%.

Table~\ref{tab:per-shoe-results} shows the grasping success rates of all methods broken down by shoe, combining the results from both the hidden- and visible-grasp viewpoints.  The full \ptgnet-\hofnet system has an overall success rate of 91\%, matching the performance of both oracle systems.  It's also worth noting that the 7/8 overall success rate on the Oxford shoe, which was not included in the training set, is comparable to the other shoes that were included in training, indicating the ability to generalize to new objects within a category.

\subsection{Experiments in simulation}

\begin{figure}
    \centering
    \includegraphics[width=0.8\columnwidth]{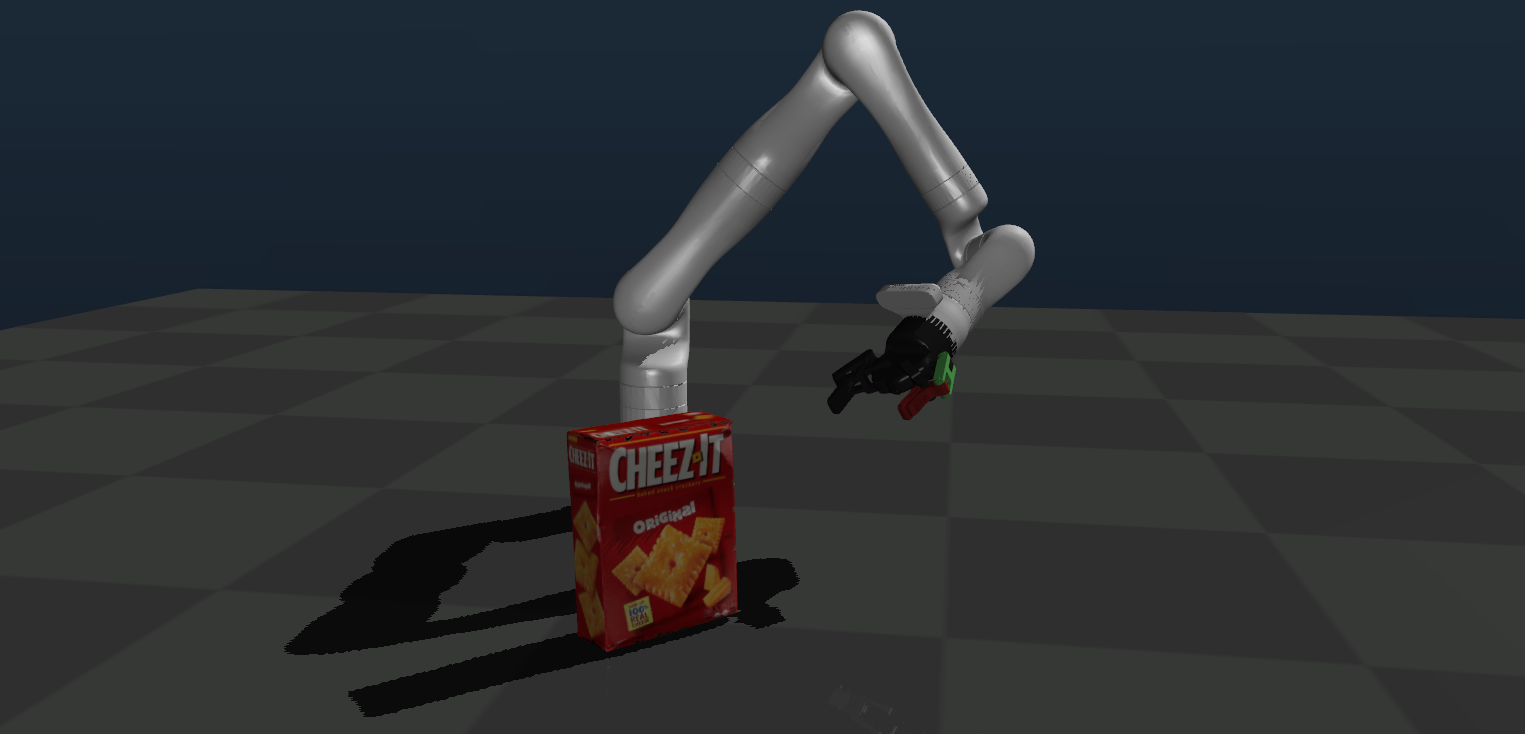}
    \caption{Simulated experimental evaluation environment similar to our physical setup - a Kinova Gen3 arm and Robotiq 2F-85 parallel-jaw gripper}
    \label{fig:mj_scene}
\end{figure}

\begin{table}
    \centering
\setlength{\tabcolsep}{5pt} 
\begin{tabular}{|c|cc|ccccc|c|}
\hline
Type     & Prop.        & Ref.          & \textbf{S} & C        & P        & G        & M        & Total    \\ \hline
Baseline & Naive        & Visible       & 73         & 60       & 88       & 75       & 30       & 70       \\
Baseline & Library      & Visible       & 93         & 78       & 98       & 98       & 81       & 92       \\
Baseline & GPD\cite{atenpas2017gpd} & - & 58         & 35       & 50       & 53       & 48       & 55       \\ \hline
Ablation & Naive        & \hofnet       & 66         & 47       & 2        & 38       & 9        & 59       \\ 
Ablation & \ptgnet      & None          & 69         & 100      & 100      & 98       & 68       & 76       \\ 
Ablation & \ptgnet      & Visible       & 79         & 81       & 98       & 95       & 63       & 81       \\ \hline
\bf{Full}& \bf{\ptgnet} & \bf{\hofnet}  & \bf{82}    & \bf{98}  & \bf{100} & \bf{98}  & \bf{68}  & \bf{84}  \\ \hline
Oracle   & \ptgnet      & Library       & 82         & 98       & 100      & 98       & 83       & 86       \\ 
Oracle   & Library      & \hofnet       & 79         & 70       & 96       & 88       & 73       & 80       \\ 
\hline
\end{tabular}
    \caption{Per-object grasping success rates (\%). C = Cracker, P = Pudding, G = Gelatin, M = Potted meat, \textbf{S} = Sugar.  The sugar box is a test object. We evaluated on 600 views of the test object and 60 views of the train objects, all using RGB-D information from the YCB dataset\cite{ycbdataset2017ijrr} and placing the object within our simulation in a kinematically feasible pose.}
    \label{tab:mj_experiments}
\end{table}

We create an experimental evaluation environment in  Mujoco \cite{todorov2012mujoco} consisting of a Kinova Gen3 arm and Robotiq 2F-85 parallel-jaw gripper, similar to our physical setup (Figure~\ref{fig:mj_scene}). We place the object in the environment such that the desired grasp is kinematically feasible, while using the RGB-D information from the BigBIRD-collected YCB dataset as inputs to \ptgnet and \hofnet. As our simulated and physical robots are utilizing a shared API and underlying control code and we are relying on real sensor data, it is expected that this simulated environment is an accurate portrayal of real-world grasping.


We train our system using 90\% of the views of our train objects --- the cracker box, gelatin box, pudding box, and potted meat --- and leave out the sugar box as a test object. We evaluate our system using all views of the the sugar box as well as the held out 10\% of the views of the train objects to highlight our methods performance on both novel objects and novel views of previously seen objects.

In addition to the previously discussed grasp proposal methods, we also evaluate the performance of the Grasp Pose Detection (GPD) algorithm~\cite{atenpas2017gpd}. The segmented depth image of the object scene is converted into a point cloud and provided as input to their open-source implementation with provided configuration. For each scene, candidate grasps are generated and the grasp with the highest score is executed.

Table \ref{tab:mj_experiments} shows in-simulation evaluation results of our method, GPD, and previously tested combinations of grasp proposal and refinement methods broken down by YCB object as well as in aggregate. The full \ptgnet-\hofnet system has an overall success rate of 91\%, 

The full \ptgnet-\hofnet system has an overall success rate of 84\%. Our system outperforms a competing grasp planning method, GPD \cite{atenpas2017gpd} which had an overall success rate of 55\%. GPD is particularly sensitive to partial views of the image, unlike our method which leverage the reconstructed point cloud, and has a high failure rate when only one or few sides of the object are visible. It is particularly worth noting that when \ptgnet is combined with various methods of refinement --- in increasing order of information present: with no refinement, with the visible point cloud, with \hofnet, and with the ground truth library point cloud --- overall performance increases. As with the on robot experiments, it is also of note that our system performs well on the sugar box, unseen at train time, with a success rate of 82\%.

\section{Discussion}
Our experimental results illustrate the value of including both a learned grasp proposer and a dedicated 3D-reconstruction component in our architecture. From convenient viewpoints of our physical setup, our grasp proposal model is able to produce successful 6-DOF grasps on its own. However, from challenging viewpoints  of our physical setup that hide the desired grasp point, 3D reconstruction boosts performance, allowing our combined system to succeed where a purely pixel-space system would fail. 

This is further reiterated by our results in simulation with YCB objects. With these solid, box-like objects, the desired grasp is no longer on the edge of the object (as on the shoe) and snapping to the nearest visible point often leads failure as the object and gripper collide. Especially for views where the desired grasp location is not visible, leveraging 3D reconstruction allows our system to refine not solely on information present within the visible depth image. 

Our system is not without limitations.  Alignment of the reconstructed point cloud to the visible point cloud via ICP is time consuming, and is sometimes a source of error, especially when very little of the target object is visible, as in pose H of the hidden-grasp experiment (Figure~\ref{fig:experimental_setup}).  In, particular, alignment is brittle when \hofnet reconstructs the shoe inaccurately (for example, generating a shape similar to a flat or sneaker in response to a challenging viewpoint of a high heel).  This is not an issue when evaluating in simulation as we have perfect pose information. In the future, we may explore a learned model to align the point clouds, which could result in better performance and faster speed.

Refining grasps by projection onto the reconstructed point cloud increases grasping performance.  However, there may be utility in performing refinement in a more sophisticated way.  Our current refinement method changes the position of the gripper, but leaves its orientation unchanged.  Refining the grasp orientation based on the shape of the nearby reconstructed surface could result in better performance - for example, by searching for an orientation where the fingertips are at antipodal points with respect to a flat surface.
\section{Conclusion and Future Work}
We presented a grasp planning system that generates accurate 6-DOF grasps from single RGB-D images of an object.  Two neural networks process the same input image to produce a 6-DOF grasp proposal and a 3D point cloud reconstruction.  By projecting the grasp onto the nearest point in the point cloud, our combined system is able to produce more accurate grasps than the grasp proposal network alone. Our experiments (on robot and in simulation) demonstrated the efficacy of our system at grasping both known and unknown objects, in addition to highlighting the benefit of combining a learned grasp proposal with 3D reconstruction.

Our current architecture uses separately trained networks for reconstruction and grasping which provides flexibility.
In the future, we hope to explore an implementation of this system where \ptgnet and \hofnet are combined into a single network with two heads and a shared convolutional encoder.  This would increase computational efficiency, and potentially result in better performance as well - training the two networks jointly could allow the loss function to capture the overall system performance (quality of the grasp post-refinement), and allow for the proposal and reconstruction heads to coordinate their outputs for better performance.

%
\bibliographystyle{IEEEtran}
\bibliography{references}


\end{document}